\documentclass[11pt,a4paper]{article}
\usepackage[hyperref]{acl2020}
\usepackage{times}
\usepackage{latexsym}

\usepackage{microtype}

\usepackage{amsfonts,amsmath,booktabs,multirow,enumerate,subcaption,makecell,graphicx}

\aclfinalcopy 

\title{An Effective Transition-based Model for Discontinuous NER}

\author{\begin{tabular}{cccc}
Xiang Dai$^{1,2}$ & Sarvnaz Karimi$^{1}$ & \textbf{Ben Hachey$^{3}$} & \textbf{Cecile Paris$^{1}$}
\end{tabular}\\
\begin{tabular}{cccc}
\multicolumn{4}{c}{$^{1}$CSIRO Data61, Sydney, Australia}\\
\multicolumn{4}{c}{$^{2}$University of Sydney, Sydney, Australia}\\
\multicolumn{4}{c}{$^{3}$Harrison.ai, Sydney, Australia}\\
\multicolumn{4}{c}{\tt \{dai.dai,sarvnaz.karimi,cecile.paris\}@csiro.au}\\
\multicolumn{4}{c}{\tt ben.hachey@gmail.com} \\
\end{tabular}
}

\date{}

\begin{document}
\maketitle

\begin{abstract}
Unlike widely used Named Entity Recognition (NER) data sets in generic domains, biomedical NER data sets often contain mentions consisting of discontinuous spans.
Conventional sequence tagging techniques encode Markov assumptions that are efficient but preclude recovery of these mentions. 
We propose a simple, effective transition-based model with generic neural encoding for discontinuous NER.
Through extensive experiments on three biomedical data sets, we show that our model can effectively recognize discontinuous mentions without sacrificing the accuracy on continuous mentions.

\end{abstract}

\section{Introduction}
\label{section:introduction}
Named Entity Recognition (NER) is a critical component of biomedical natural language processing applications. In pharmacovigilance, it can be used to identify adverse drug events in consumer reviews in online medication forums, alerting medication developers, regulators and clinicians~\cite{Leaman:Wojtulewicz:BioNLP:2010,Sarker:Ginn:JBI:2015,Karimi:Wang:Survey:2015}.
In clinical settings, NER can be used to extract and summarize key information from electronic medical records such as conditions hidden in unstructured doctors' notes~\cite{Feblowitz:Wright:JBI:2011,Wang:Wang:JBI:2018}. These applications require identification of complex mentions not seen in generic domains~\cite{Dai:ACL-SRW:2018}.

\begin{figure}[t]
    \centering
    \begin{subfigure}[b]{0.4\textwidth}
        \includegraphics[width=0.8\textwidth]{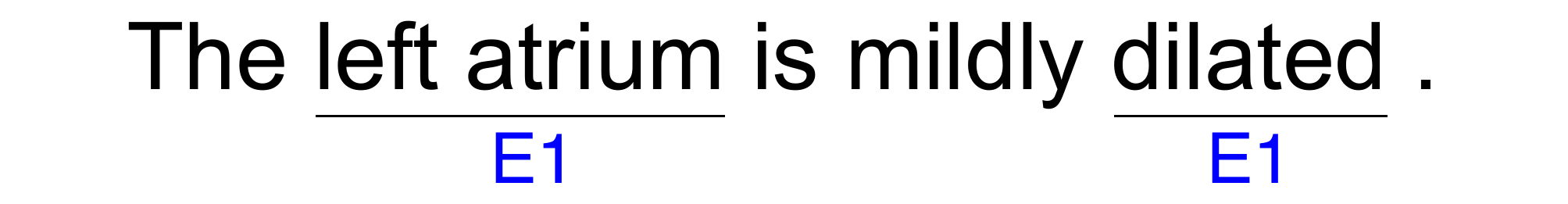}
    \end{subfigure}
    \\
    \vspace{1em}
    \begin{subfigure}[b]{0.4\textwidth}
        \includegraphics[width=0.8\textwidth]{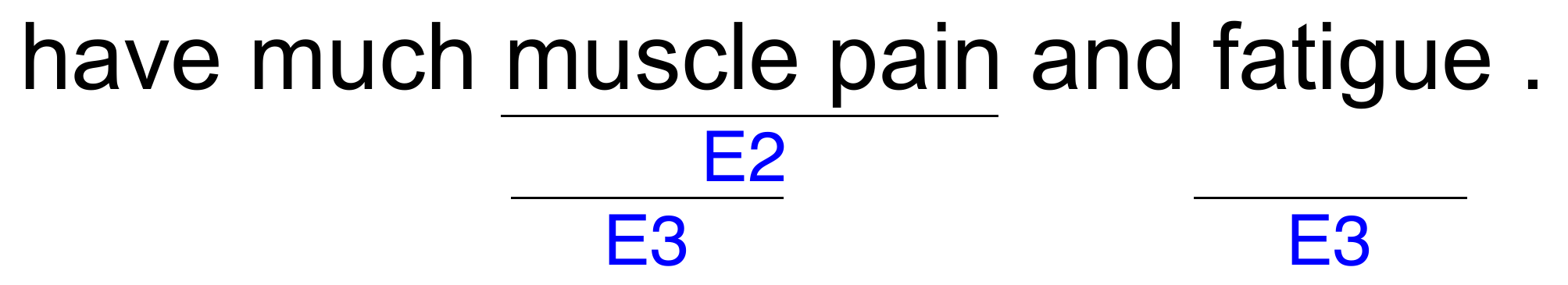}
    \end{subfigure}
    \caption{Examples involving discontinuous mentions, taken from the ShARe 13~\citep{Pradhan:Elhadad:CLEF:2013} and CADEC~\citep{Karimi:Metke:JBI:2015} data sets, respectively. The first example contains a discontinuous mention \emph{`left atrium dilated'}, the second example contains two mentions that overlap: \emph{`muscle pain'} and \emph{`muscle fatigue'} (discontinuous).~\label{figure1-example}}
\end{figure}

Widely used sequence tagging techniques ({\em flat model}) encode two assumptions that do not always hold: (1) mentions do not nest or overlap, therefore each token can belong to at most one mention; and, (2) mentions comprise continuous sequences of tokens.
Nested entity recognition addresses violations of the first assumption~\cite{Lu:Roth:EMNLP:2015,Katiyar:Cardie:NAACL:2018,Sohrab:Miwa:EMNLP:2018,Ringland:Dai:ACL:2019}.
However, the violation of the second assumption is comparatively less studied and requires handling discontinuous mentions (see examples in Figure~\ref{figure1-example}).


In contrast to continuous mentions which are often short spans of text, {\em discontinuous mentions} consist of \textbf{components} that are separated by \textbf{intervals}.
Recognizing discontinuous mentions is particularly challenging as exhaustive enumeration of possible mentions, including discontinuous and overlapping spans, is exponential in sentence length.
Existing approaches for discontinuous NER either suffer from high time complexity~\cite{McDonald:Crammer:EMNLP:2005} or ambiguity in translating intermediate representations into mentions~\cite{Tang:Cao:BMC:2013,Metke-Jimenez:Karimi:BMDID:2016,Muis:Lu:EMNLP:2016}.
In addition, current art uses traditional approaches that rely on manually designed features, which are tailored to recognize specific entity types. 
Also, these features usually do not generalize well in different genres~\cite{Leaman:Khare:JBI:2015}.

\paragraph{Motivations}
The main motivation for recognizing discontinuous mentions is that they usually represent {\em compositional concepts} that differ from concepts represented by individual components. 
For example, the mention \textit{`left atrium dilated'} in the first example of Figure~\ref{figure1-example} describes a disorder which has its own CUI (Concept Unique Identifier) in UMLS (Unified Medical Language System), whereas both \textit{`left atrium'} and \textit{`dilated'} also have their own CUIs. We argue that, in downstream applications such as pharmacovigilance and summarization, recognizing these discontinuous mentions that refer to disorders or symptoms is more useful than recognizing separate components which may refer to body locations or general feelings.

Another important characteristic of discontinuous mentions is that they usually {\em overlap}. That is, several mentions may share components that refer to the same body location (e.g., \textit{`muscle'} in \textit{`muscle pain and fatigue'}), or the same feeling (e.g., \textit{`Pain'} in \textit{`Pain in knee and foot'}). Separating these overlapping mentions rather than identifying them as a single mention is important for downstream tasks, such as entity linking where the assumption is that the input mention refers to one entity~\cite{Shen:Wang:TKDE:2015}.

\paragraph{Contributions}
We propose an end-to-end transition-based model with generic neural encoding that allows us to leverage specialized actions and attention mechanism to determine whether a span is the component of a discontinuous mention or not.\footnote{Code available at GitHub: \href{https://github.com/daixiangau/acl2020-transition-discontinuous-ner}{https://bit.ly/2XazEAO}}
We evaluate our model on three biomedical data sets with a substantial number of discontinuous mentions and demonstrate that our model can effectively recognize discontinuous mentions without sacrificing the accuracy on continuous mentions.

\section{Prior Work}
Existing methods on discontinuous NER can be mainly categorized into two categories: token level approach, based on sequence tagging techniques, and sentence level approach, where a combination of mentions within a sentence is jointly predicted~\citep{Dai:ACL-SRW:2018}.

\paragraph{Token level approach}
Sequence tagging model takes a sequence of tokens as input and outputs a tag for each token, composed of a position indicator (e.g., BIO schema) and an entity type.
The vanilla BIO schema cannot effectively represent discontinuous, overlapping mentions, therefore, some studies overcome this limitation via expanding the BIO tag set~\cite{Tang:Cao:BMC:2013,Metke-Jimenez:Karimi:BMDID:2016,Dai:Karimi:ALTA:2017,Tang:Hu:Wireless:2018}. 
In addition to BIO indicators, four new position indicators are introduced in~\cite{Metke-Jimenez:Karimi:BMDID:2016} to represent discontinuous mentions that may overlap:
\begin{itemize}
    \item \textbf{BH}: \textbf{B}eginning of \textbf{H}ead, defined as the components shared by multiple mentions;
    \item \textbf{IH}: \textbf{I}ntermediate of \textbf{H}ead;
    \item \textbf{BD}: \textbf{B}eginning of \textbf{D}iscontinuous body, defined as the exclusive components of a discontinuous mention; and
    \item \textbf{ID}: \textbf{I}ntermediate of \textbf{D}iscontinuous body.
\end{itemize}

\paragraph{Sentence level approach}
Instead of predicting whether each token belongs to an entity mention and its role in the mention, sentence level approach predicts a combination of mentions within a sentence. 
A hypergraph, proposed by \citeauthor{Lu:Roth:EMNLP:2015} (2015) and extended in~\cite{Muis:Lu:EMNLP:2016}, can compactly represent discontinuous and overlapping mentions in one sentence. 
A sub-hypergraph of the complete hypergraph can, therefore, be used to represent a combination of mentions in the sentence. 
For the token at each position, there can be six different node types:
\begin{itemize}
    \item \textbf{A}: mentions that start from the current token or a future token;
    \item \textbf{E}: mentions that start from the current token;
    \item \textbf{T}: mentions of a certain entity type that start from the current token;
    \item \textbf{B}: mentions that contain the current token;
    \item \textbf{O}: mentions that have an interval at the current token;
    \item \textbf{X}: mentions that end at the current token.
\end{itemize}
Using this representation, a single entity mention can be represented as a path from node \textbf{A} to node \textbf{X}, incorporating at least one node of type \textbf{B}.

%

Note that both token level and sentence level approaches predict first an intermediate representation of mentions (e.g., a sequence of tags in~\cite{Metke-Jimenez:Karimi:BMDID:2016} and a sub-hypergraph in~\cite{Muis:Lu:EMNLP:2016}), which are then decoded into the final mentions. During the final decoding stage, both models suffer from some level of ambiguity. Taking the sequence tagging model using BIO variant schema as an example, even if the model can correctly predict the gold sequence of tags for the example sentence `muscle pain and fatigue' (BH I O BD), it is still not clear whether the token \textit{`muscle'} forms a mention by itself, because the same sentence containing three mentions (\textit{`muscle'}, \textit{`muscle pain'} and \textit{`muscle fatigue'}) can be encoded using the same gold sequence of tags. We refer to a survey by~\cite{Dai:ACL-SRW:2018} for more discussions on these models, and~\cite{Muis:Lu:EMNLP:2016} for a theoretical analysis of ambiguity of these models.


%

Similar to prior work, our proposed transition-based model uses an intermediate representation (i.e., a sequence of actions). However, it does not suffer from this ambiguity issue. That is, the output sequence of actions can always be unambiguously decoded into mention outputs.

The other two methods that focus on the discontinuous NER problem in literature are described in~\citep{McDonald:Crammer:EMNLP:2005,Wang:Lu:EMNLP:2019}. \citet{McDonald:Crammer:EMNLP:2005} solve the NER task as a structured multi-label classification problem. 
Instead of starting and ending indices, they represent each entity mention using the set of token positions that belong to the mention. 
This representation is flexible, as it allows mentions consisting of discontinuous tokens and does not require mentions to exclude each other. However, this method suffers from high time complexity. \citet{Tang:Hu:Wireless:2018} compare this representation with BIO variant schema proposed in~\cite{Metke-Jimenez:Karimi:BMDID:2016}, and found that they achieve competitive $F_1$ scores, although the latter method is more efficient. A two-stage approach that first detects all components and then combines components into discontinuous mentions based on a classifier's decision was explored in recent work by~\citet{Wang:Lu:EMNLP:2019}.

\begin{figure}[t]
    \centering
    \includegraphics[width=0.3\textwidth]{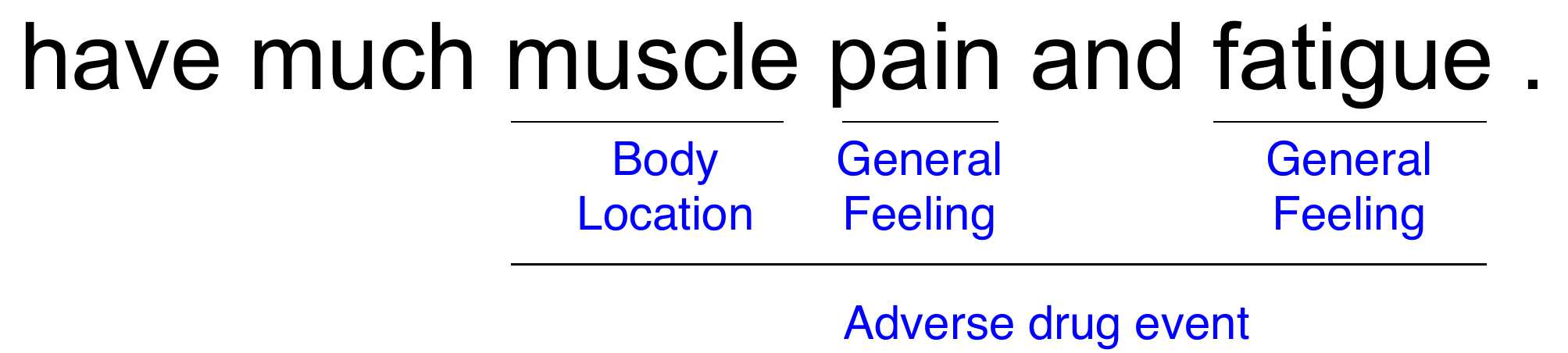}
    \caption{Examples involving Nested mentions.~\label{figure1-example-nested-ner}}
\end{figure}

\paragraph{Discontinuous NER vs. Nested NER}
Although discontinuous mentions may overlap, we discriminate this overlapping from the one in nested NER. That is, if one mention is completely contained by the other, we call mentions involved nested entity mentions. In contrast, overlapping in discontinuous NER is usually that two mentions overlap, but no one is completely contained by the other. Most of existing nested NER models are built to tackle the complete containing structure~\citep{Finkel:Manning:EMNLP:2009,Lu:Roth:EMNLP:2015}, and they cannot be directly used to identify overlapping mentions studied in this paper, nor mention the discontinuous mentions. However, we note that there is a possible perspective to solve discontinuous NER task by adding fine-grained entity types into the schema. Taking the second sentence in Figure~\ref{figure1-example} as an example, we can add two new entity types: `Body Location' and 'General Feeling', and then annotate `muscle pain and fatigue' as a `Adverse drug event' mention, `muscle' as a `Body Location' mention, and `pain' and `fatigue' as `General Feeling' mentions (Figure~\ref{figure1-example-nested-ner}). Then the discontinuous NER task can be converted into a Nested NER task.

\section{Model}

Transition-based models, due to their high efficiency, are widely used for NLP tasks, such as parsing and entity recognition~\cite{Chen:Manning:EMNLP:2014,Lample:Ballesteros:NAACL:2016,Lou:Zhang:BioInfo:2017,Wang:Lu:EMNLP:2018-transition}. The model we propose for discontinuous NER is based on the shift-reduce parser~\cite{Watanabe:Sumita:ACL:2015,Lample:Ballesteros:NAACL:2016} that employs a \textbf{stack} to store partially processed spans and a \textbf{buffer} to store unprocessed tokens. The learning problem is then framed as: given the state of the parser, predict an action which is applied to change the state of the parser. This process is repeated until the parser reaches the end state (i.e., the stack and buffer are both empty).

\begin{figure}[t]
    \centering
    \includegraphics[width=1\linewidth]{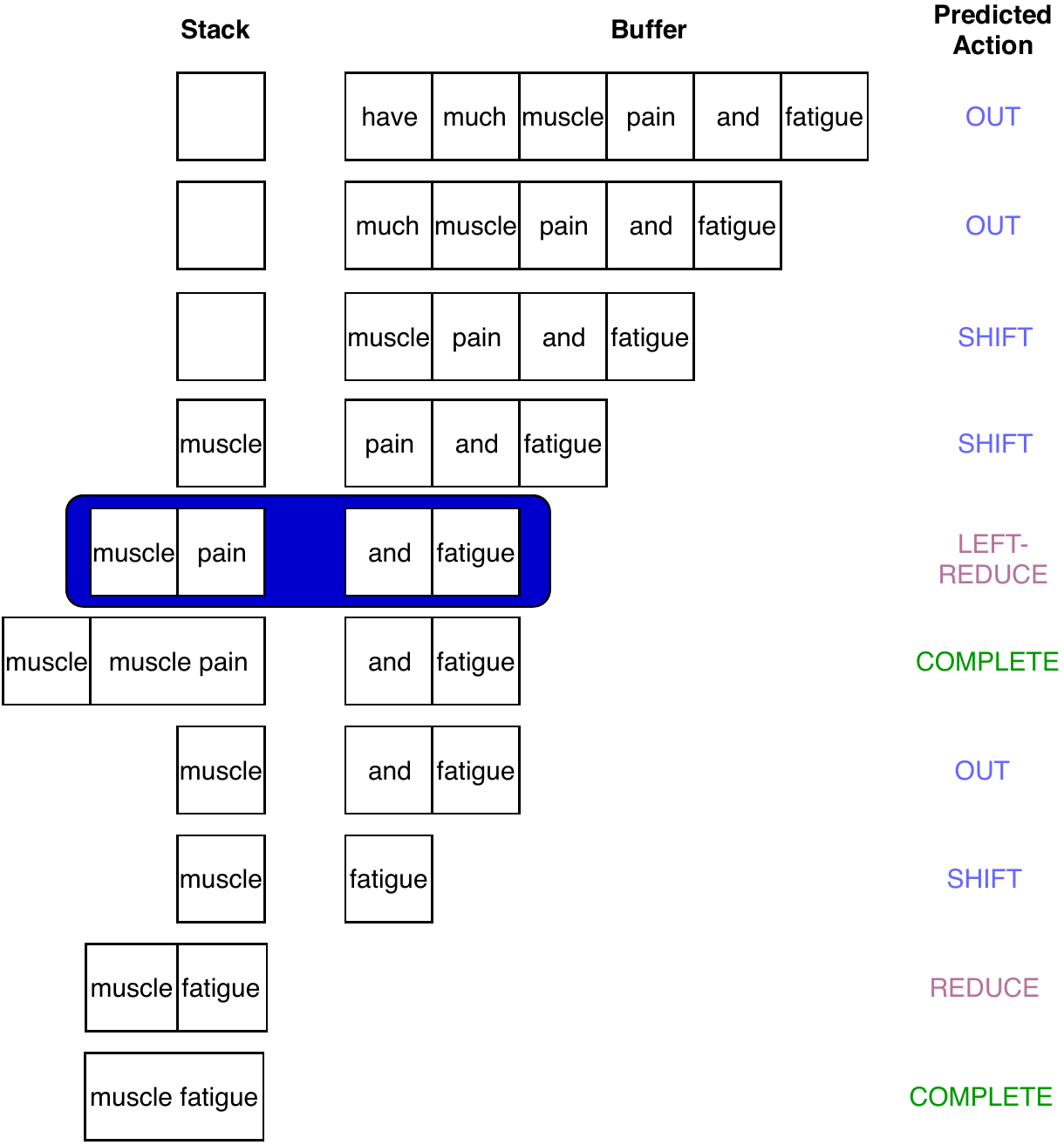}
    \caption{An example sequence of transitions. Given the states of stack and buffer (blue highlighted), as well as the previous actions, predict the next action (i.e., LEFT-REDUCE) which is then applied to change the states of stack and buffer.}
    \label{figure2-transition-example}
\end{figure}


The main difference between our model and the ones in \citep{Watanabe:Sumita:ACL:2015,Lample:Ballesteros:NAACL:2016} is the set of transition actions. \citet{Watanabe:Sumita:ACL:2015} use SHIFT, REDUCE, UNARY, FINISH, and IDEA for the constituent parsing system. \citet{Lample:Ballesteros:NAACL:2016} use SHIFT, REDUCE, OUT for the flat NER system. Inspired by these models, we design a set of actions specifically for recognizing discontinuous and overlapping structure. There are in total six actions in our model:
\begin{itemize}
    \item SHIFT moves the first token from the buffer to the stack; it implies this token is part of an entity mention. 
    \item OUT pops the first token of the buffer, indicating it does not belong to any mention. 
    \item COMPLETE pops the top span of the stack, outputting it as an entity mention. If we are interested in multiple entity types, we can extend this action to COMPLETE-$y$ which labels the mention with entity type $y$.
    \item REDUCE pops the top two spans $s_0$ and $s_1$ from the stack and concatenates them as a new span which is then pushed back to the stack.
    \item LEFT-REDUCE is similar to the REDUCE action, except that the span $s_1$ is kept in the stack. This action indicates the span $s_1$ is involved in multiple mentions. In other words, several mentions share $s_1$ which could be a single token or several tokens.
    \item RIGHT-REDUCE is the same as LEFT-REDUCE, except that $s_0$ is kept in the stack.
\end{itemize}

Figure~\ref{figure2-transition-example} shows an example about how the parser recognizes entity mentions from a sentence. 
Note that, given one parser state, not all types of actions are valid. 
For example, if the stack does not contain any span, only SHIFT and OUT actions are valid because all other actions involve popping spans from the stack. We employ hard constraints that we only select the most likely action from valid actions.

\subsection{Representation of the Parser State}
\label{section:configuration-representation}

Given a sequence of $N$ tokens, we first run a bi-directional LSTM~\citep{Graves:Mohamed:ICASSP:2013} to derive the contextual representation of each token. Specifically, for the $i$-th token in the sequence, its representation can be denoted as:
\[
\mathbf{\tilde{c_i}} = \left [ \overrightarrow{\text{LSTM}} (\mathbf{t_0}, \dots, \mathbf{t_{i}}); \overleftarrow{\text{LSTM}} (\mathbf{t_{i}}, \dots, \mathbf{t_{N - 1}}) \right ],
\]
where $\mathbf{t_i}$ is the concatenation of the embeddings for the $i$-th token, its character level representation learned using a CNN network~\cite{Ma:Hovy:ACL:2016}. Pretrained contextual word representations have shown its usefulness on improving various NLP tasks. Here, we can also concatenate pretrained contextual word representations using ELMo~\cite{Peters:Neumann:NAACL:2018} with $\mathbf{\tilde{c_i}}$, resulting in:
\begin{equation}
    \label{equ:elmo}
    \mathbf{c_i} = \left [ \mathbf{\tilde{c_i}}; \text{ELMo}_i \right ],
\end{equation}
where $\text{ELMo}_i$ is the output representation of pretrained ELMo models (frozen) for the $i$-th token.
These token representations $\mathbf{c}$ are directly used to represent tokens in the buffer. We also explore a variant that uses the output of pretrained BERT~\citep{Devlin:Chang:NAACL:2019} as token representations $\mathbf{c}$, and fine-tune the BERT model. However, this fine-tuning approach with BERT does not achieve as good performance as feature extraction approach with ELMo~\cite{Peters:Ruder:arXiv:2019}. 

Following the work in~\cite{Dyer:Ballesteros:ACL:2015}, we use Stack-LSTM to represent spans in the stack. That is, if a token is moved from the buffer to the stack, its representation is learned using:
\[
\mathbf{s_0} = \text{Stack-LSTM} (\mathbf{s_D} \dots \mathbf{s_1};\mathbf{c}_{\text{SHIFT}}),
\]
where $D$ is the number of spans in the stack. Once REDUCE related actions are applied, we use a multi-layer perceptron to learn the representation of the concatenated span. For example, the REDUCE action takes the representation of the top two spans in the stack: $\mathbf{s_0}$ and $\mathbf{s_1}$, and produces a new span representation:
\[
\mathbf{\Tilde{s}} = \textbf{W}^T [\mathbf{s_0}; \mathbf{s_1}] + b,
\]
where $\textbf{W}$ and $b$ denote the parameters for the composition function. 
The new span representation $\mathbf{\Tilde{s}}$ is pushed back to the stack to replace the original two spans: $\mathbf{s_0}$ and $\mathbf{s_1}$.

\subsection{Capturing Discontinuous Dependencies}
We hypothesize that the interactions between spans in the stack and tokens in the buffer are important factors in recognizing discontinuous mentions. 
Considering the example in Figure~\ref{figure2-transition-example}, a span in the stack (e.g., \textit{`muscle'}) may need to combine with a future token in the buffer (e.g., \textit{`fatigue'}). To capture this interaction, we use multiplicative attention~\cite{Luong:Pham:ACL:2015} to let the span in the stack $\mathbf{s_i}$ learn which token in the buffer to attend, and thus a weighted sum of the representation of tokens in the buffer $\mathbf{B}$:

\begin{equation}
\mathbf{s^a_i} = \textbf{softmax} (\mathbf{s_i^TW^a_iB})\mathbf{B}.
\label{equ:attn}
\end{equation}
We use distinct $\mathbf{W_i^a}$ for $\mathbf{s_i}$ separately.

\subsection{Selecting an Action}
Finally, we build the parser representation as the concatenation of the representation of top three spans from the stack ($\mathbf{s_0}, \mathbf{s_1}, \mathbf{s_2}$) and its attended representation ($\mathbf{s^a_0}$, $\mathbf{s^a_1}$, $\mathbf{s^a_2}$), as well as the representation of the previous action $\mathbf{a}$, which is learned using a simple unidirectional LSTM. If there are less than 3 spans in the stack or no previous action, we use randomly initialized vectors $\mathbf{s_{empty}}$ or $\mathbf{a_{empty}}$ to replace the corresponding vector. This parser representation is used as input for the final softmax prediction layer to select the next action.

\section{Data sets}
Although some text annotation tools, such as BRAT~\cite{Stenetorp:Pyysalo:EACL:2012}, allow discontinuous annotations, corpora annotated with a large number of discontinuous mentions are still rare.
We use three data sets from the biomedical domain: CADEC~\cite{Karimi:Metke:JBI:2015}, ShARe 13~\cite{Pradhan:Elhadad:CLEF:2013} and ShARe 14~\cite{Mowery:Velupillai:CLEF:2014}.
Around 10\% of mentions in these three data sets are discontinuous. 
The descriptive statistics are listed in Table~\ref{table2-data-statistics}. 

\begin{table}[t]
\centering
\begin{small}
\setlength{\tabcolsep}{2pt} 
\begin{tabular}{r  c  c  c}
\toprule
& \bf CADEC & \bf ShARe 13 & \bf ShARe 14 \\ \midrule
Text type & online posts & clinical notes & clinical notes \\ 
Entity type & ADE & Disorder & Disorder \\ 
\# Documents & 1,250 & 298 & 433 \\ 
\# Tokens & 121K & 264K & 494K \\ 
\# Sentences & 7,597 & 18,767 & 34,618 \\
\# Mentions & 6,318 & 11,161 & 19,131 \\
\# Disc.M & 675 (10.6) & 1,090 (9.7) & 1,710 (8.9) \\ 
\midrule
Avg mention L. & 2.7 & 1.8 & 1.7 \\
Avg Disc.M L. & 3.5 & 2.6 & 2.5 \\
Avg interval L. & 3.3 & 3.0 & 3.2 \\ 
\midrule
\multicolumn{4}{c}{\bf Discontinuous Mentions} \\
\midrule
2 components & 650 (95.7) & 1,026 (94.3) & 1,574 (95.3) \\ 
3 components & \phantom{0}27 (\phantom{0}3.9) & \phantom{00}62 (\phantom{0}5.6) & \phantom{00}76 (\phantom{0}4.6) \\ 
4 components & \phantom{00}2 (\phantom{0}0.2) & \phantom{000}0 (\phantom{0}0.0) & \phantom{000}0 (\phantom{0}0.0) \\
\midrule
No overlap & \phantom{0}82 (12.0) & \phantom{0}582 (53.4) & \phantom{0}820 (49.6) \\
Overlap at left & 351 (51.6) & \phantom{0}376 (34.5) & \phantom{0}616 (37.3) \\
Overlap at right & 152 (22.3) & \phantom{0}102 (\phantom{0}9.3) & \phantom{0}170 (10.3) \\
Multiple overlaps & \phantom{0}94 (13.8) & \phantom{00}28 (\phantom{0}2.5) & \phantom{00}44 (\phantom{0}2.6) \\
\midrule
\multicolumn{4}{c}{\bf Continuous Mentions} \\
\midrule
Overlap & 326 (\phantom{0}5.7) & \phantom{0}157 (\phantom{0}1.5) & \phantom{0}228 (\phantom{0}1.3) \\
\bottomrule
\end{tabular}
\caption{The descriptive statistics of the data sets. ADE: adverse drug events; Disc.M: discontinuous mentions; Disc.M L.: discontinuous mention length, where intervals are not counted. Numbers in parentheses are the percentage of each category.~\label{table2-data-statistics}}
\end{small}
\end{table}


CADEC is sourced from \textit{AskaPatient}\footnote{https://www.askapatient.com/}, a forum where patients can discuss their experiences with medications. 
The entity types in CADEC include drug, Adverse Drug Event (ADE), disease and symptom.
We only use ADE annotations because only the ADEs involve discontinuous annotations. 
This also allows us to compare our results directly against previously reported results~\cite{Metke-Jimenez:Karimi:BMDID:2016,Tang:Hu:Wireless:2018}. 
ShARe 13 and 14 focus on the identification of disorder mentions in clinical notes, including discharge summaries, electrocardiogram, echocardiogram, and radiology reports~\cite{Johnson:Pollard:SD:2016}. 
A disorder mention is defined as any span of text which can be mapped to a concept in the disorder semantic group of SNOMED-CT~\cite{Cornet:Keizer:BMC:2008}.

Although these three data sets share similar field (the subject matter of the content being discussed), the tenor (the participants in the discourse, their relationships to each other, and their purposes) of CADEC is very different from the ShARe data sets~\cite{Dai:Karimi:NAACL:2019}. 
In general, laymen (i.e., in CADEC) tend to use idioms to describe their feelings, whereas professional practitioners (i.e., in ShARe) tend to use compact terms for efficient communications. 
This also results in different features of discontinuous mentions between these data sets, which we will discuss further in $\S$~\ref{section-analysis}.


\paragraph{Experimental Setup}
As CADEC does not have an official train-test split, we follow~\citet{Metke-Jimenez:Karimi:BMDID:2016} and randomly assign 70\% of the posts as the training set, 15\% as the development set, and the remaining posts as the test set.
\footnote{These splits can be downloaded from \href{https://github.com/daixiangau/acl2020-transition-discontinuous-ner}{https://bit.ly/2XazEAO}.} 
The train-test splits of ShARe 13 and 14 are both from their corresponding shared task settings, except that we randomly select 10\% of documents from each training set as the development set. 
Micro average strict match $F_1$ score is used to evaluate the effectiveness of the model. 
The trained model which is most effective on the development set, measured using the $F_1$ score, is used to evaluate the test set. 

\section{Baseline Models}

\begin{table*}[t]
\begin{small}
    \centering
    \begin{tabular}{cr  cccc  cccc  cccc}
    \toprule
    &&& \multicolumn{3}{c}{\bf CADEC} && \multicolumn{3}{c}{\bf ShARe 13} && \multicolumn{3}{c}{\bf ShARe 14} \\ 
    \cmidrule{4-6}\cmidrule{8-10} \cmidrule{12-14}
    &\bf Model && P & R & F && P & R & F && P & R & F \\ 
    \cmidrule{2-2} \cmidrule{4-6}\cmidrule{8-10} \cmidrule{12-14}
    & \cite{Metke-Jimenez:Karimi:BMDID:2016} && 64.4 & 56.5 & 60.2 && -- & -- & -- && -- & -- & -- \\
    & \cite{Tang:Hu:Wireless:2018} && 67.8 & 64.9 & 66.3 && -- & -- & -- && -- & -- & -- \\
    & \cite{Tang:Wu:CLEF:2014} && -- & -- & -- && 80.0 & 70.6 & 75.0 && -- & -- & -- \\
    \hline
    & Flat && 65.3 & 58.5 & 61.8 && 78.5 & 66.6 & 72.0 && 76.2 & 76.7 & 76.5 \\ 
    & BIO Extension && 68.7 & 66.1 & 67.4 && 77.0 & 72.9 & 74.9 && 74.9 & 78.5 & 76.6 \\ 
    & Graph && \bf 72.1 & 48.4 & 58.0 && \bf 83.9  & 60.4  & 70.3 && \bf 79.1 & 70.7 & 74.7 \\ 
    & Ours && 68.9 & \bf 69.0 & \bf 69.0 && 80.5 & \bf 75.0 & \bf 77.7 && 78.1 & \bf 81.2 & \bf 79.6 \\ 
    \bottomrule 
    \end{tabular}
    \caption{Evaluation results on the whole test set in terms of precision, recall and $F_1$ score. The original ShARe 14 task focuses on template filling of disorder attributes: that is, given a disorder mention, recognize the attribute from its context. In this work, we use its mention annotations and frame the task as a discontinuous NER task.~\label{table3-main-result}}
\end{small}
\end{table*}
\begin{table*}
\begin{small}
\setlength{\tabcolsep}{2.5pt} 
    \centering
    \begin{tabular}{cr  cccc  cccc  cccc  cccc  cccc  cccc}
    \toprule
    &&& \multicolumn{11}{c}{\bf Sentences with discontinuous mentions} && \multicolumn{11}{c}{\bf Discontinuous mentions only} \\ 
    \cmidrule{4-14}\cmidrule{16-26}
    &&& \multicolumn{3}{c}{\bf CADEC} && \multicolumn{3}{c}{\bf ShARe 13} && \multicolumn{3}{c}{\bf ShARe 14} && \multicolumn{3}{c}{\bf CADEC} && \multicolumn{3}{c}{\bf ShARe 13} && \multicolumn{3}{c}{\bf ShARe 14} \\ 
    \cmidrule{4-6}\cmidrule{8-10} \cmidrule{12-14} \cmidrule{16-18}\cmidrule{20-22} \cmidrule{24-26}
    & \bf Model && P & R & F && P & R & F && P & R & F && P & R & F && P & R & F && P & R & F \\ 
    \cmidrule{2-2} \cmidrule{4-6}\cmidrule{8-10} \cmidrule{12-14} \cmidrule{16-18}\cmidrule{20-22} \cmidrule{24-26}
    & Flat && 50.2 & 36.7 & 42.4 && 43.5 & 28.1 & 34.2 && 41.5 & 31.9 & 36.0 && 0 & 0 & 0 && 0 & 0 & 0 && 0 & 0 & 0 \\ 
    & BIO E. && 63.8 & 52.0 & 57.3 && 51.8 & 39.5 & 44.8 && 37.5 & 38.4 & 37.9 && 5.8 & 1.0 & 1.8 && 39.7 & 12.3 & 18.8 && 8.8 & 4.5 & 6.0 \\ 
    & Graph && \bf 69.5 & 43.2 & 53.3 && \bf 82.3 & 47.4 & 60.2 && 60.0 & 52.8 & 56.2 && \bf 60.8 & 14.8 & 23.9 && 78.4 & 36.6 & 50.0 && 42.7 & 39.5 & 41.1 \\ 
    & Ours && 66.5 & \bf 64.3 & \bf 65.4 && 70.5 & \bf 56.8 & \bf 62.9 && \bf 61.9 & \bf 64.5 & \bf 63.1 && 41.2 & \bf 35.1 & \bf 37.9 && \bf 78.5 & \bf 39.4 & \bf 52.5 && \bf 56.1 & \bf 43.8 & \bf 49.2 \\ 
    \bottomrule 
    \end{tabular}
    \caption{Evaluation results on sentences that contain at least one discontinuous mention (left part) and on discontinuous mentions only (right part).~\label{table3-result-on-disc}}
\end{small}
\end{table*}

We choose one flat NER model which is strong at recognizing continuous mentions, and two discontinuous NER models as our baseline models:

\paragraph{Flat model}
To train the flat model on our data sets, we use an off-the-shelf framework: Flair~\cite{Akbik:Blythe:COLING:2018}, which achieves the state-of-the-art performance on CoNLL 03 data set.
Recall that the flat model cannot be directly applied to data sets containing discontinuous mentions. 
Following the practice in~\cite{Stanovsky:Gruhl:EACL:2017}, we replace the discontinuous mention with the shortest span that fully covers it, and merge overlapping mentions into a single mention that covers both. 
Note that, different from~\cite{Stanovsky:Gruhl:EACL:2017}, we apply these changes only on the training set, but not on the development set and the test set.

\paragraph{BIO extension model}
The original implementation in~\cite{Metke-Jimenez:Karimi:BMDID:2016} used a CRF model with manually designed features. 
We report their results on CADEC in Table~\ref{table3-main-result} and re-implement a BiLSTM-CRF-ELMo model using their tag schema (denoted as `BIO Extension' in Table~\ref{table3-main-result}).

\paragraph{Graph-based model}
The original paper of~\cite{Muis:Lu:EMNLP:2016} only reported the evaluation results on sentences which contain at least one discontinuous mention. 
We use their implementation to train the model and report evaluation results on the whole test set (denoted as `Graph' in Table~\ref{table3-main-result}). 
We argue that it is important to see how a discontinuous NER model works not only on the discontinuous mentions but also on all the mentions, especially since, in real data sets, the ratio of discontinuous mentions cannot be made a priori.

We do not choose the model proposed in~\cite{Wang:Lu:EMNLP:2019} as the baseline model, because it is based on a strong assumption about the ratio of discontinuous mentions. \citet{Wang:Lu:EMNLP:2019} train and evaluate their model on sentences that contain at least one discontinuous mention. Our early experiments show that the effectiveness of their model strongly depends on this assumption. In contrast, we train and evaluate our model in a more practical setting where the number of continuous mentions is much larger than the one of discontinuous mentions.

\section{Experimental Results}
When evaluated on the whole test set, our model outperforms three baseline models, as well as over previous reported results in the literature, in terms of recall and $F_1$ scores (Table~\ref{table3-main-result}). 

The graph-based model achieves highest precision, but with substantially lower recall, therefore obtaining lowest $F_1$ scores.
In contrast, our model improves recall over flat and BIO extension models as well as previously reported results, without sacrificing precision. 
This results in more balanced precision and recall. 
Improved recall is especially encouraging for our motivating pharmacovigilance and medical record summarization applications, where recall is at least as important as precision.


\paragraph{Effectiveness on recognizing discontinuous mentions}
Recall that only 10\% of mentions in these three data sets are discontinuous. 
To evaluate the effectiveness of our proposed model on recognizing discontinuous mentions, we follow the evaluation approach in~\cite{Muis:Lu:EMNLP:2016} where we construct a subset of test set where only sentences with at least one discontinuous mention are included (Left part of Table~\ref{table3-result-on-disc}). 
We also report the evaluation results when only discontinuous mentions are considered (Right part of Table~\ref{table3-result-on-disc}).
Note that sentences in the former setting usually contain continuous mentions as well, including those involved in overlapping structure (e.g., `muscle pain' in the sentence `muscle pain and fatigue'). 
Therefore, the flat model, which cannot predict any discontinuous mentions, still achieves 38\% $F_1$ on average when evaluated on these sentences with at least one discontinuous mention, but 0\% $F_1$ when evaluated on discontinuous mentions only.

Our model again achieves the highest $F_1$ and recall in all three data sets under both settings. The comparison between these two evaluation results also shows the necessity of comprehensive evaluation settings. 
The BIO E. model outperforms the graph-based model in terms of $F_1$ score on CADEC, when evaluated on sentences with discontinuous mentions. 
However, it achieves only 1.8 $F_1$ when evaluated on discontinuous mentions only. 
The main reason is that most of discontinuous mentions in CADEC are involved in overlapping structure (88\%, cf. Table~\ref{table2-data-statistics}), and the BIO E. model is better than the graph-based model at recognizing these continuous mentions. 
On ShARe 13 and 14, where the portion of discontinuous mentions involved in overlapping is much less than on CADEC, the graph-based model clearly outperforms BIO E. model in both evaluation settings.


\section{Analysis}
\label{section-analysis}

We start our analysis from characterizing discontinuous mentions from the three data sets. 
Then we measure the behaviors of our model and two discontinuous NER models on the development sets based on characteristics identified and attempt to draw conclusions from these measurements. 

\subsection{Characteristics of Discontinuous Mentions}
\label{subsection-characteristics}

\begin{figure*}[t]
\begin{subfigure}{.32\textwidth}
  \centering
  \includegraphics[width=.98\linewidth]{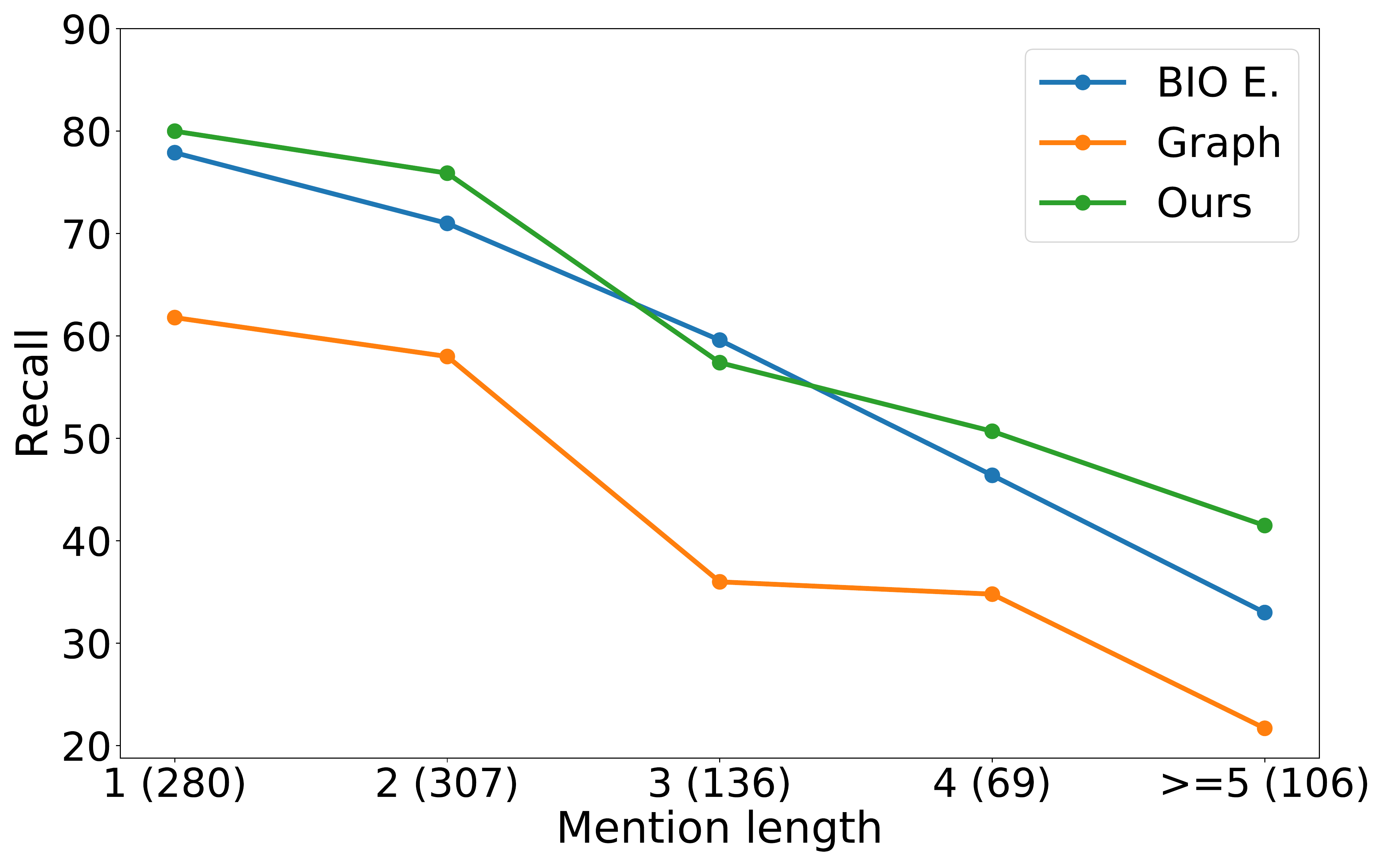}
  \caption{CADEC}
\end{subfigure}%
\begin{subfigure}{.32\textwidth}
  \centering
  \includegraphics[width=.98\linewidth]{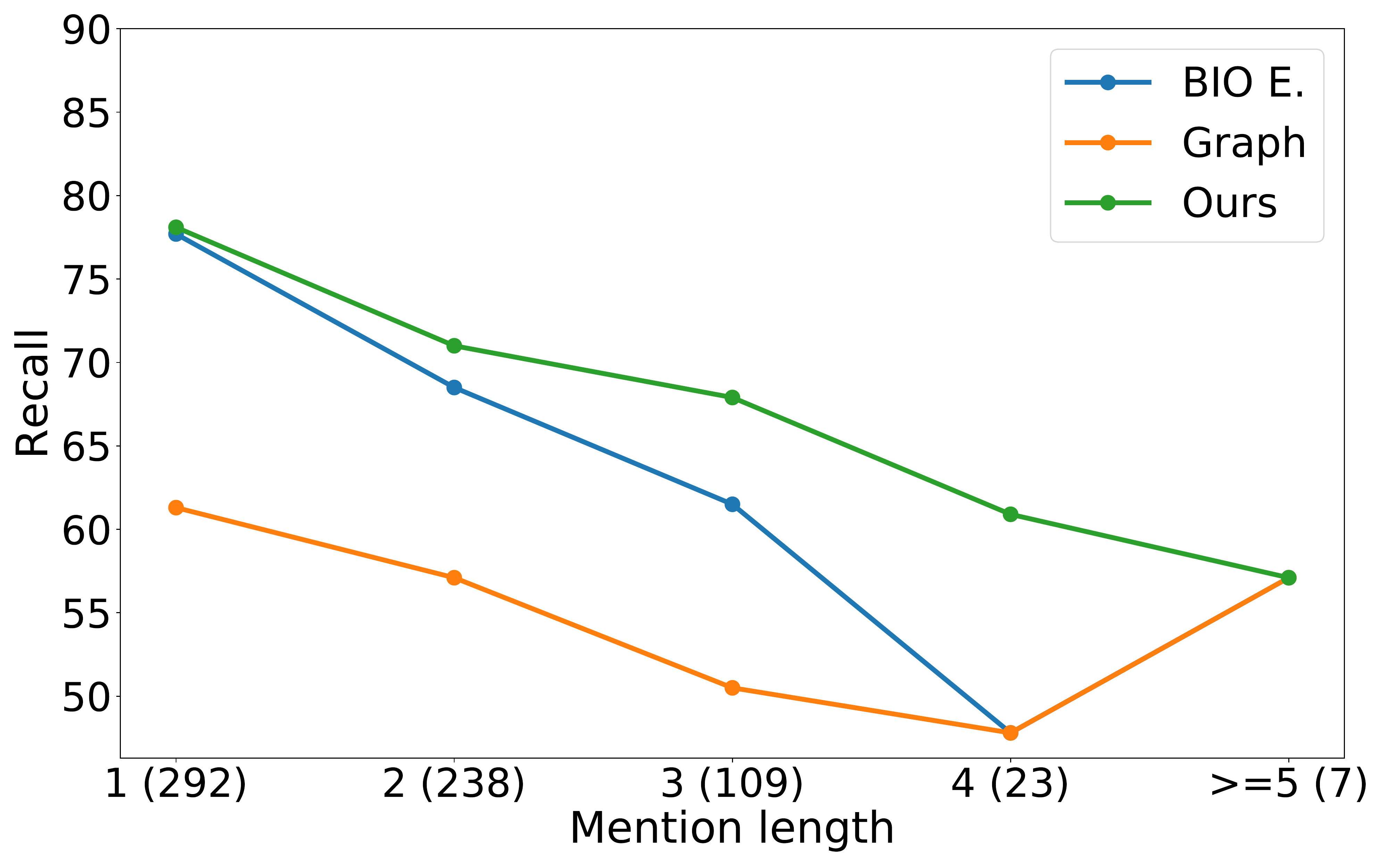}
  \caption{ShARe 13}
\end{subfigure}%
\begin{subfigure}{.32\textwidth}
  \centering
  \includegraphics[width=.98\linewidth]{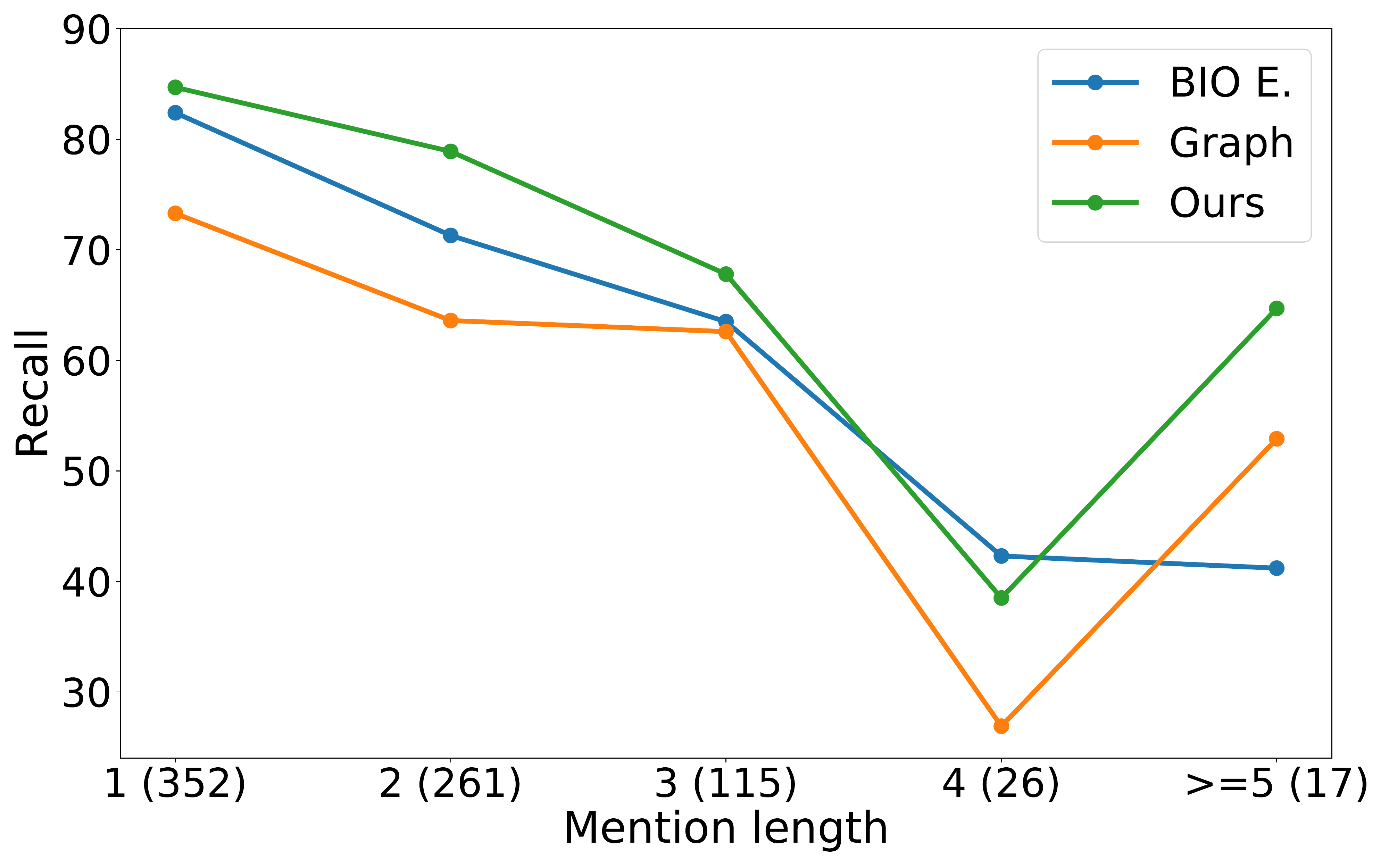}
  \caption{ShARe 14}
  \label{fig:sfig1}
\end{subfigure}

\begin{subfigure}{.32\textwidth}
  \centering
  \includegraphics[width=.98\linewidth]{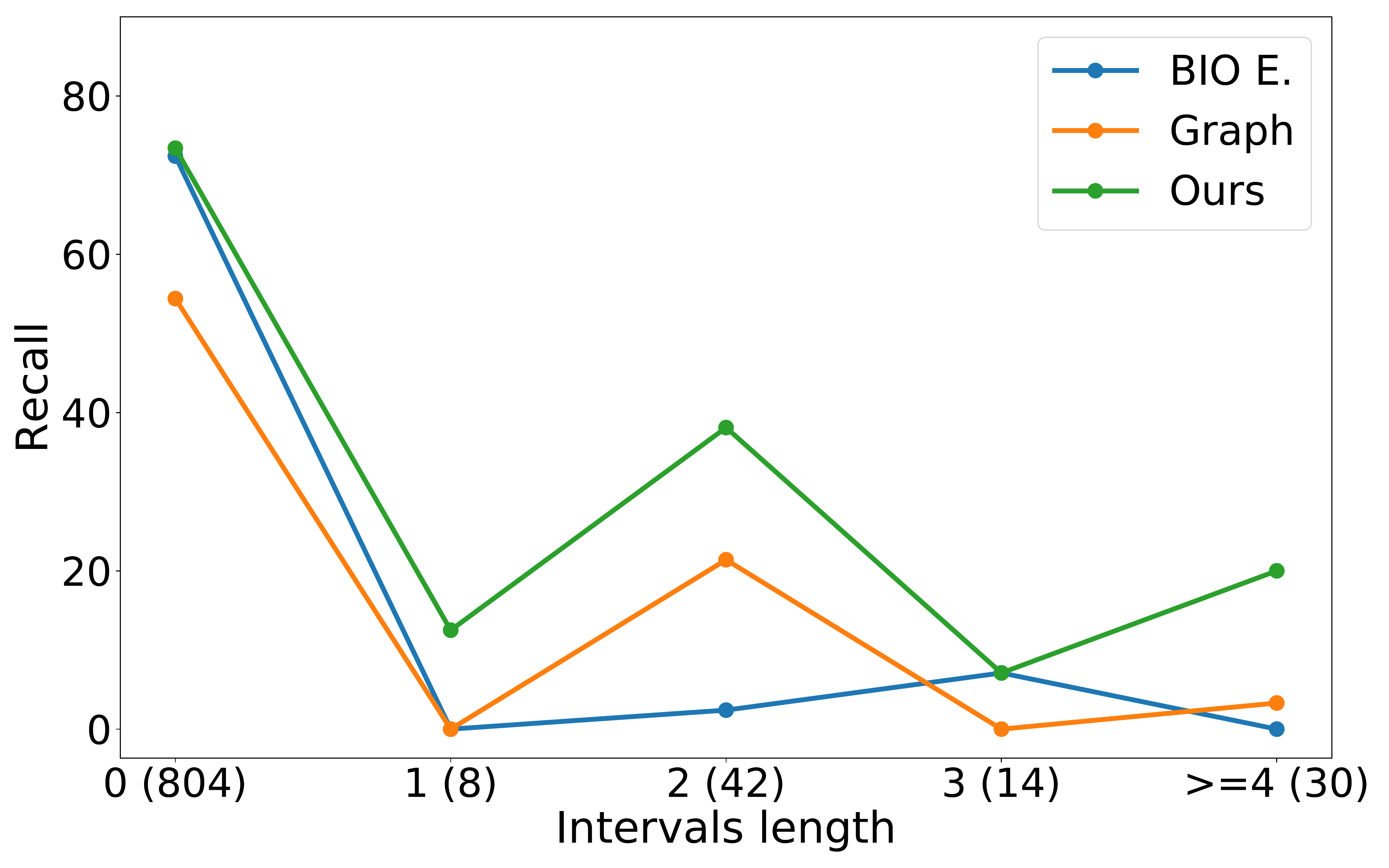}
  \caption{CADEC}
\end{subfigure}%
\begin{subfigure}{.32\textwidth}
  \centering
  \includegraphics[width=.98\linewidth]{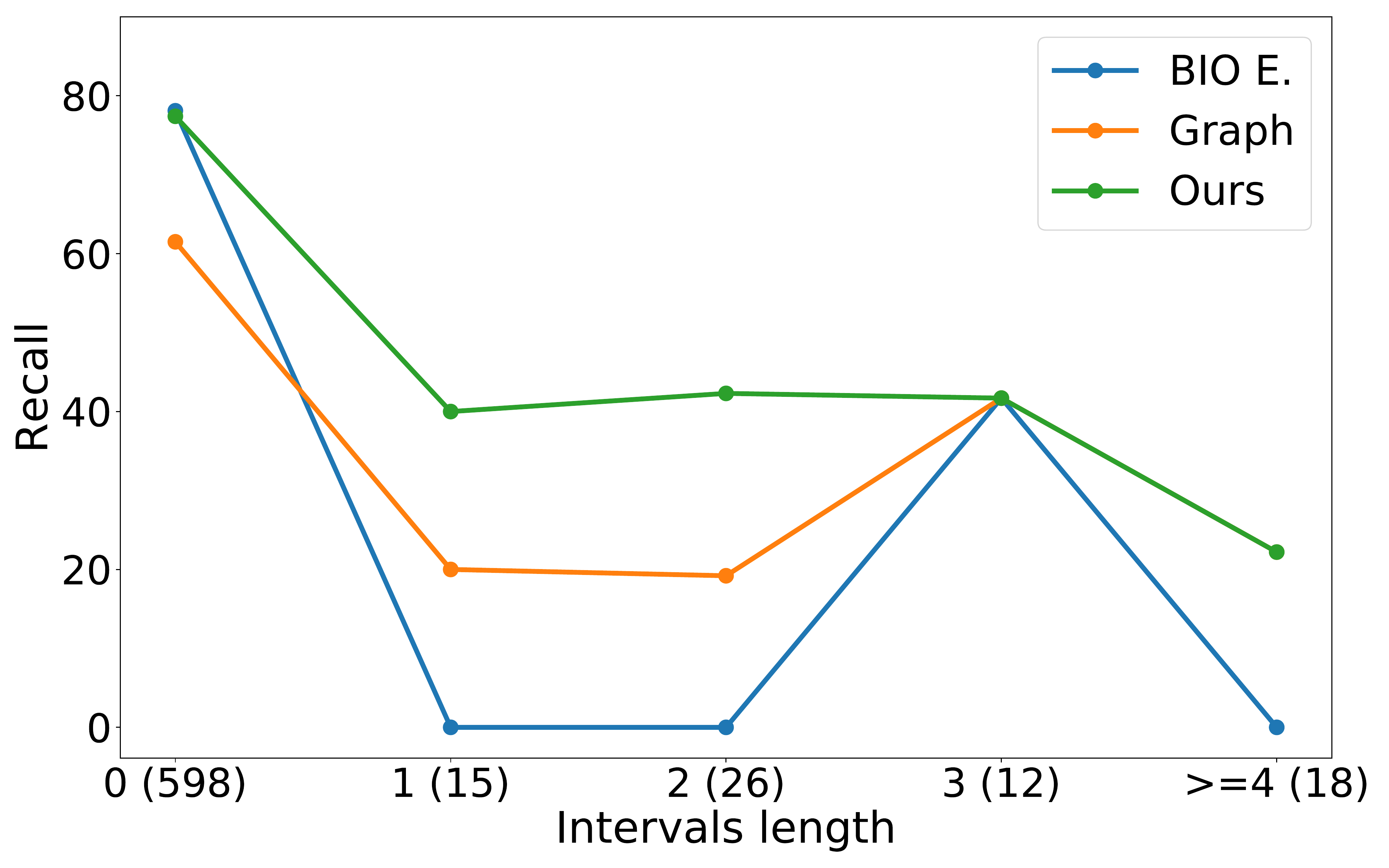}
  \caption{ShARe 13}
  \label{fig:sfig1}
\end{subfigure}%
\begin{subfigure}{.32\textwidth}
  \centering
  \includegraphics[width=.98\linewidth]{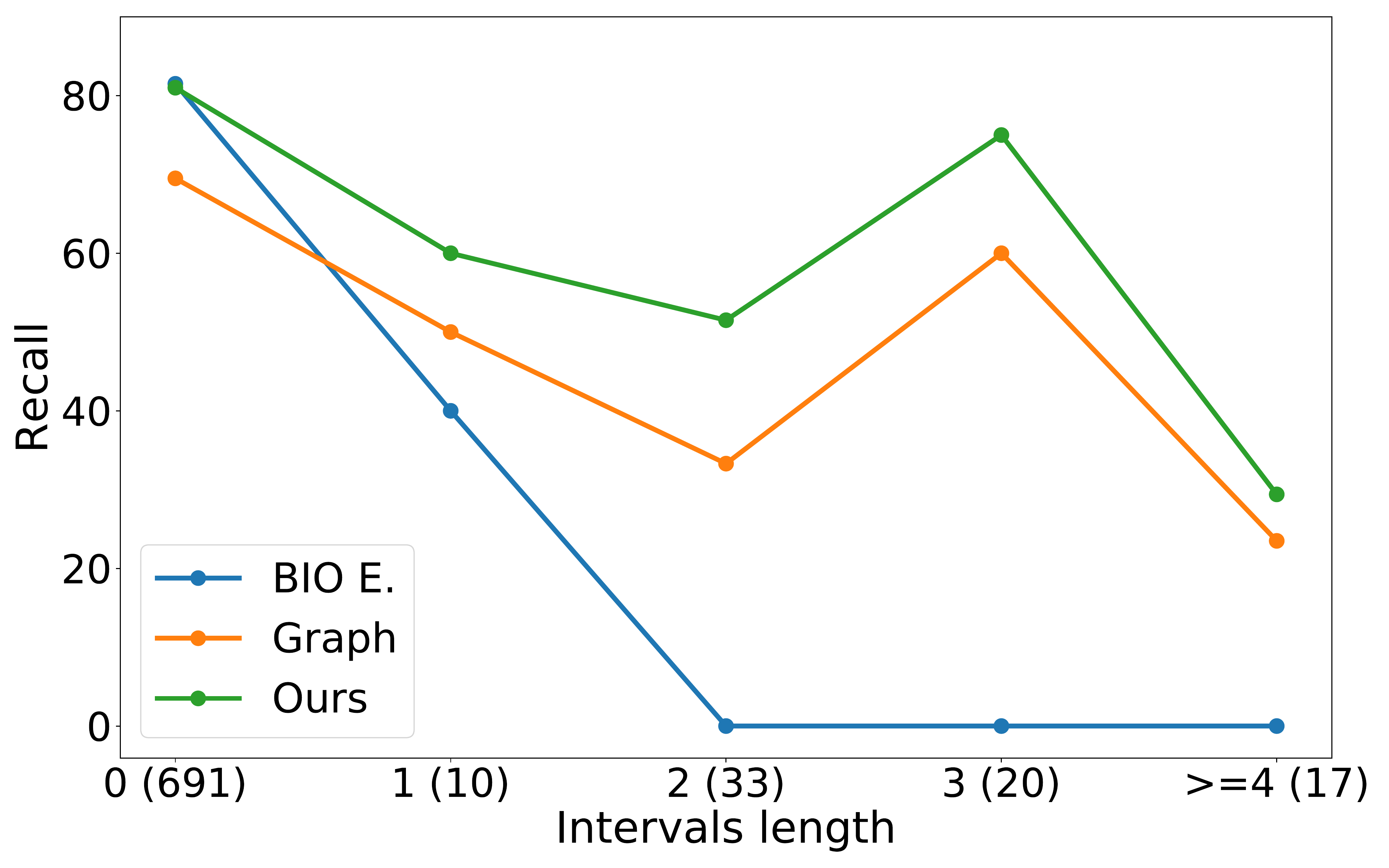}
  \caption{ShARe 14}
\end{subfigure}
\caption{The impact of mention length and interval length on recall. Mentions with interval length of zero are continuous mentions. Numbers in parentheses are the number of gold mentions.~\label{figure3-length-vs-recall}}
\end{figure*}
Recall that discontinuous mentions usually represent compositional concepts that consist of multiple components. 
Therefore, discontinuous mentions are usually longer than continuous mentions (Table~\ref{table2-data-statistics}). 
In addition, intervals between components make the total length of span involved even longer. 
Previous work shows that flat NER performance degrades when applied on long mentions~\cite{Augenstein:Das:SemEval:2017,Xu:Jiang:ACL:2017}.

Another characteristic of discontinuous mentions is that they usually {\em overlap} (cf. $\S$~\ref{section:introduction}). 
From this perspective, we can categorize discontinuous mentions into four categories:
\begin{itemize}
    \item No overlap: in such cases, the discontinuous mention can be intervened by severity indicators (e.g., \textit{`is  mildly'} in sentence \textit{`left atrium is mildly dilated'}), preposition (e.g., \textit{`on my'} in sentence \textit{`...rough on my stomach...'}) and so on. 
    This category accounts for half of discontinuous mentions in the ShARe data sets but only 12\% in CADEC (Table~\ref{table2-data-statistics}).
    \item Left overlap: the discontinuous mention shares one component with other mentions, and the shared component is at the beginning of the discontinuous mention. 
    This is usually accompanied with coordination structure (e.g., the shared component \textit{`muscle'} in \textit{`muscle pain and fatigue'}). 
    Conjunctions (e.g., \textit{`and'}, \textit{`or'}) are clear indicators of the coordination structure. 
    However, clinical notes are usually written by practitioners under time pressure.
    They often use commas or slashes rather than conjunctions.
    This category accounts for more than half of discontinuous mentions in CADEC and one third in ShARe.
    \item Right overlap: similar to left overlap, although the shared component is at the end. 
    For example, \textit{`hip/leg/foot pain'} contains three mentions that share \textit{`pain'}. 
    \item Multi-overlap: the discontinuous mention shares multiple components with the others, which usually forms \emph{crossing compositions}. 
    For example, the sentence \textit{`Joint and Muscle Pain / Stiffness'} contains four mentions: \textit{`Joint Pain'}, \textit{`Joint Stiffness'}, \textit{`Muscle Stiffness'} and \textit{`Muscle Pain'}, where each discontinuous mention share two components with the others.
\end{itemize}


\subsection{Impact of Overlapping Structure}
Previous study shows that the intervals between components can be problematic for coordination boundary detection~\cite{Ficler:Goldberg:EMNLP:2016}. 
Conversely, we want to observe whether the overlapping structure may help or hinder discontinuous entity recognition.
We categorize discontinuous mentions into different subsets, described in $\S$~\ref{subsection-characteristics}, and measure the effectiveness of different discontinuous NER models on each category. 

From Table~\ref{table-impact-of-overlapping}, we find that our model achieves better results on discontinuous mentions belonging to `No overlap' category on ShARe 13 and 14, and `Left overlap' category on CADEC and ShARe 14.
Note that `No overlap' category accounts for half of discontinuous mentions in ShARe 13 and 14, whereas `Left overlap' accounts for half in CADEC (Table~\ref{table2-data-statistics}). 
Graph-based model achieves better results on `Right overlap' category.
On the `Multi-overlap' category, no models is effective, which emphasizes the challenges of dealing with this syntactic phenomena. 
We note, however, the portion of discontinuous mentions belonging to this category is very small in all three data sets.

Although our model achieves better results on `No overlap' category on ShARe 13 and 14, it does not predict correctly any discontinuous mention belonging to this category on CADEC.
The ineffectiveness of our model, as well as other discontinuous NER models, on CADEC `No overlap' category can be attributed to two reasons: 1) the number of discontinuous mentions belonging to this category in CADEC is small (around 12\%), rending the learning process more difficult.
2) the gold annotations belonging to this category are inconsistent from a linguistic perspective. 
For example, severity indicators are annotated as the interval of the discontinuous mention sometimes, but not often. 
Note that this may be reasonable from a medical perspective, as some symptoms are roughly grouped together no matter their severity, whereas some symptoms are linked to different concepts based on their severity.

\begin{table}[t]
\begin{small}
\setlength{\tabcolsep}{3pt} 
    \centering
    \begin{tabular}{ c r c  cc c cc c cc}
    \toprule
    & & & \multicolumn{2}{c}{\bf CADEC} && \multicolumn{2}{c}{\bf ShARe 13} && \multicolumn{2}{c}{\bf ShARe 14} \\
    \cmidrule{4-5}\cmidrule{7-8}\cmidrule{10-11}
     & \bf Model & & \# & F && \# & F && \# & F \\
     \cmidrule{2-2}\cmidrule{4-5}\cmidrule{7-8}\cmidrule{10-11}
     
    \multirow{3}{*}{No $\mathbb{O}$} & BIO E. && \multirow{3}{*}{9} & 0.0 && \multirow{3}{*}{41} & 7.5 && \multirow{3}{*}{39} & 0.0 \\
    & Graph && & 0.0 &&  & 32.1 &&  & 45.2 \\
    & Ours &&  & 0.0 && & \bf 36.1 && & \bf 57.1 \\
    \hline
    
    \multirow{3}{*}{Left $\mathbb{O}$} & BIO E. && \multirow{3}{*}{54} & 6.0 && \multirow{3}{*}{11} & 25.0 && \multirow{3}{*}{30} & 15.7 \\
    & Graph && & 9.2 &&  & \bf 45.5 &&  & 37.7 \\
    & Ours &&  & \bf 28.6 && & 33.3 && & \bf 49.2 \\
    \hline
    
    \multirow{3}{*}{Right $\mathbb{O}$} & BIO E. && \multirow{3}{*}{16} & 0.0 && \multirow{3}{*}{19} & 0.0 && \multirow{3}{*}{5} & 0.0 \\
    & Graph && & \bf 45.2 &&  & \bf 21.4 &&  & 0.0 \\
    & Ours &&  & 29.3 && & 13.3 &&  & 0.0 \\
    \hline
    
    \multirow{3}{*}{Multi $\mathbb{O}$} & BIO E. && \multirow{3}{*}{15} & 0.0 && \multirow{3}{*}{0} & -- && \multirow{3}{*}{6} & 0.0 \\
    & Graph && & 0.0 &&  & -- &&  & 0.0 \\
    & Ours &&  & 0.0 && & -- && & 0.0 \\
    \bottomrule
    \end{tabular}
    \caption{Evaluation results on different categories of discontinuous mentions. `\#' columns show the number of gold discontinuous mentions in development set of each category. $\mathbb{O}$: overlap.\label{table-impact-of-overlapping}}
\end{small}
\end{table}

\subsection{Impact of Mention and Interval Length}
We conduct experiments to measure the ability of different models on recalling mentions of different lengths, and to observe the impact of interval lengths. 
We found that the recall of all models decreases with the increase of mention length in general (Figure~\ref{figure3-length-vs-recall} (a -- c)), which is similar to previous observations in the literature on flat mentions. 
However, the impact of interval length is not straightforward. 
Mentions with very short interval lengths are as difficult as those with very long interval lengths to be recognized (Figure~\ref{figure3-length-vs-recall} (d -- f)). 
On CADEC, discontinuous mentions with interval length of 2 are easiest to be recognized (Figure~\ref{figure3-length-vs-recall} (d)), whereas those with interval length of 3 are easiest on ShARe 13 and 14. We hypothesize this also relates to annotation inconsistency, because very short intervals may be overlooked by annotators.

In terms of model comparison, our model achieves highest recall in most settings. 
This demonstrates our model is effective to recognize both continuous and discontinuous mentions with various lengths. 
In contrast, the BIO E. model is only strong at recalling continuous mentions (outperforming the graph-based model), but fails on discontinuous mentions (interval lengths $>$ 0).

\subsection{Example Predictions}
We find that previous models often fail to identify discontinuous mentions that involve long and overlapping spans. For example, the sentence ‘Severe joint pain in the shoulders and knees.’ contains two mentions: ‘Severe joint pain in the shoulders’ and ‘Severe joint pain in the knees’. Graph-based model does not identify any mention from this sentence, resulting in a low recall. The BIO extension model predicts most of  these tags (8 out of 9) correctly, but fails to decode into correct mentions (predict ‘Severe joint pain in the’, resulting in a false positive, while it misses ‘Severe joint pain in the shoulders’). In contrast, our model correctly identifies both of these two mentions.

No model can fully recognize mentions which form crossing compositions. For example, the sentence ‘Joint and Muscle Pain / Stiffness’ contains four mentions: ‘Joint Pain’, ‘Joint Stiffness’, ‘Muscle Stiffness’ and ‘Muscle Pain’, all of which share multiple components with the others. Our model correctly predicts ‘Joint Pain’ and ‘Muscle Pain’, but it mistakenly predicts ‘Stiffness’ itself as a mention.

\section{Summary}

We propose a simple, effective transition-based model that can recognize discontinuous mentions without sacrificing the accuracy on continuous mentions. We evaluate our model on three biomedical data sets with a substantial number of discontinuous mentions. Comparing against two existing discontinuous NER models, our model is more effective, especially in terms of recall.

\section*{Acknowledgments}
We would like to thank Danielle Mowery for helping us to obtain the ShARe data sets. We also thank anonymous reviewers for their insightful comments. Xiang Dai is supported by Sydney University's Engineering and Information Technologies Research Scholarship as well as CSIRO's Data61 top up scholarship. 

\bibliography{references}
\bibliographystyle{acl_natbib}

\end{document}